\documentclass{article}

 \usepackage[preprint]{neurips_2025}

\usepackage[utf8]{inputenc} 
\usepackage[T1]{fontenc}    
\usepackage{hyperref}       
\usepackage{url}            
\usepackage{booktabs}       
\usepackage{amsfonts}       
\usepackage{subcaption}
\usepackage{nicefrac}       
\usepackage{microtype}      
\usepackage{xcolor}         
\usepackage{amsmath}
\usepackage{tikz}
\usetikzlibrary{positioning, shapes.geometric}
\usepackage{algpseudocode,algorithm}
\usepackage{comment}
\title{\textbf{Vision Tiny Recursion Model (ViTRM): \\ Parameter-Efficient Image Classification via Recursive State Refinement}}

\author{
  Ange-Clément Akazan$^{1,2}$\thanks{Equal contribution.} \quad
  Abdoulaye Koroko$^{3}$\footnotemark[1] \quad
  Verlon Roel Mbingui$^{1,2}$\footnotemark[1] \\
  \textbf{Choukouriyah Arinloye}$^{1,2}$ \quad
  \textbf{Hassan Fifen}$^{1,2}$ \quad
  \textbf{Rose Bandolo}$^{1,2}$ \\
  $\Sigma \eta iGm\alpha$ Research Group$^{1}$, AIMS RIC$^{2}$, SaH Analytics International$^{3}$ \\
  \texttt{\small \{aakazan, vmbingui, carinloye, hfifen, ressomba\}@aimsric.org} \\
  \texttt{\small abdoulaye.koroko@sahanalytics.com}
}

\begin{document}

\maketitle

\begin{abstract}
The success of deep learning in computer vision has been driven by models of
increasing scale, from deep Convolutional Neural Networks (CNN) to large Vision
Transformers (ViT). While effective, these architectures are
parameter-intensive and demand significant computational resources, limiting deployment in resource-constrained environments. Inspired
by Tiny Recursive Models (TRM), which show that small recursive networks can
solve complex reasoning tasks through iterative state refinement, we introduce
the \textbf{Vision Tiny Recursion Model (ViTRM)}: a parameter-efficient
architecture that replaces the $L$-layer ViT encoder with a single tiny
$k$-layer block ($k{=}3$) applied recursively $N$ times. Despite using up to $6 \times $ and $84 \times$ fewer parameters than CNN based models and ViT respectively, ViTRM maintains competitive performance on CIFAR-10 and CIFAR-100. This demonstrates that recursive computation is a viable, parameter-efficient alternative to architectural
depth in vision.
\end{abstract}

\section{Introduction}
\label{sec1}
The field of computer vision has undergone two major architectural transformations over the past decade. The first was the dominance of Convolutional Neural Networks (CNN) \cite{Lecun1998,NIPS2012_c399862d,simonyan2014very}, exemplified by deep residual architectures such as ResNet \citep{he2016deep}. CNN owe their success to carefully designed inductive biases, including locality, weight sharing, and translation equivariance, which align well with the statistical structure of natural images. 

The second transformation emerged with the introduction of the Vision Transformer (ViT) \citep{dosovitskiy2021imageworth16x16words}, which demonstrated that convolutional priors are not strictly necessary. By treating an image as a sequence of patches and modeling global dependencies through self-attention, ViTs achieved competitive, and eventually superior, performance compared to CNN. This shift marked a move away from handcrafted spatial inductive biases toward more generic, attention-based architectures.

A defining characteristic of this progress, however, has been scale. State-of-the-art ViT models rely heavily on massive pre-training datasets such as JFT-300M and contain hundreds of millions to billions of parameters. This “deeper and larger is better” paradigm has led to impressive performance gains, but at the cost of substantial computational requirements. Training, fine-tuning, and deployment become increasingly impractical in resource-constrained environments, limiting accessibility and broader applicability. In parallel to these developments in perception, a distinct paradigm has emerged in the domain of algorithmic reasoning. Recent works such as the Hierarchical Reasoning Model (HRM) \citep{wang2025hierarchicalreasoningmodel} and the Tiny Recursive Model (TRM) \citep{jolicoeurmartineau2025recursive} demonstrate that complex reasoning tasks, such as Sudoku solving or ARC-AGI challenges, can be addressed using extremely small neural networks. Notably, TRM employs an encoder transformer-like  block  with single two-layer network applied recursively to iteratively refine both predictions and latent representations. Rather than increasing architectural depth, these models leverage recursive computation over time. This suggests an alternative scaling principle: trading parameter count for iterative refinement.

These two trajectories, massive-scale transformers for perception and tiny recursive networks for reasoning, highlight a fundamental question: can recursive computation serve as a viable alternative to architectural scaling in vision models? In other words, can the parameter-efficiency observed in recursive reasoning be transferred to perceptual tasks such as image classification?

We happily answer this in the affirmative by proposing the Vision Tiny Recursion Model (ViTRM). Our contributions are threefold:

\begin{itemize}
\item We introduce ViTRM, a novel architecture that couples the patch-based embedding of ViT with a tiny, weight-sharing recursive encoder inspired by TRM.
   
\item We experimentally demonstrate that ViTRM achieves accuracy on par with ResNet and ViT baselines while using up to 6-84× fewer parameters. 
 \item We perform an ablation study demonstrating that high number of recursion does not necessarily  lead to high accuracy. We provide insight on rational recursion and supervision steps to prefer.
\end{itemize}

The remainder of this work is organized as follows. Section~\eqref{sec2} reviews related work on vision transformers and iterative refinement. Section~\eqref{sec3} introduces the proposed ViTRM framework. Section~\eqref{sec4} presents the experimental results and their analysis.
\section{Related work}
\label{sec2}
ViTRM is positioned at the convergence of multiple research streams, including Vision Transformers and attention-free alternatives, iterative refinement, object-centric models with recurrent attention, deep equilibrium and fixed-point networks, and recursive reasoning architectures. We synthesize these threads to highlight the conceptual and practical gaps that ViTRM is designed to address.

\subsection{Transformers in Vision}

Transformers have become a central architecture in computer vision. The seminal Vision Transformer (ViT) \cite{dosovitskiy2021imageworth16x16words} demonstrated that standard Transformer architectures \cite{vaswani2017attention}, when pretrained on sufficiently large datasets, can match or exceed convolutional networks on image recognition tasks. 
However, ViT's quadratic complexity in the number of tokens and its reliance on large-scale pretraining posed practical challenges. 
DeiT \cite{touvron2021training} introduced distillation techniques that enable training competitive vision transformers on smaller datasets like ImageNet-1K without extensive pretraining. Swin Transformer \cite{liu2021swin} tackled computational efficiency by introducing hierarchical architectures with shifted windows.

These architectural innovations made transformers more practical for dense prediction tasks such as object detection and segmentation. 
A parallel research direction has examined whether self-attention is inherently required for effective modeling in computer vision. MLP-Mixer \cite{tolstikhin2021mlp} replaced self-attention with simple MLPs applied separately along spatial and channel dimensions, achieving competitive performance on image classification. 
This attention-free approach illustrated an efficiency that comes with reduced flexibility compared to attention-based methods.

\subsection{Iterative and Recursive Refinement in Vision}

The core idea of iterative refinement is to progressively improve predictions through multiple processing stages, each refining the output based on previous estimates. Detection Transformers (DETR) \cite{carion2020end} have a decoder that applies multiple layers of self- and cross-attention to refine a fixed set of learned object queries. 
Deformable DETR \cite{zhu2021deformable} builds upon this framework by replacing global attention with deformable attention, allowing each query to attend to 
a learned set of spatial locations, thereby improving performance on small objects. Cascade R-CNN \cite{cai2018cascade} adopted a different iterative strategy by sequentially training detection heads at progressively higher IoU thresholds. Similarly, Mask R-CNN \cite{he2017mask} extended detection to instance segmentation by adding a mask prediction head operating on region-of-interest features.

A key limitation across these approaches is the fixed number of refinement steps determined at training time. Unlike adaptive methods that adjust computation based on task difficulty, these models apply the same processing depth regardless of input complexity. To overcome these constraints, ViTRM introduces dynamic halting and weight-shared recursion to achieve both strong parameter efficiency and flexible test-time computation.

\subsection{Object-Centric Latents and Iterative Attention}

Another research direction focuses on learning object-centric representations through iterative attention mechanisms that decompose scenes into structured latent variables.

Slot Attention \cite{locatello2020object} iteratively refines a fixed set of latent slots through competitive attention over image features. However, the learned slots primarily capture scene decomposition and are not directly optimized for task prediction. The Perceiver \cite{jaegle2021perceiver} extends this paradigm to arbitrary input modalities using iterative cross-attention. While its refinement process resembles recursive reasoning, latent updates are not explicitly coupled with progressive task prediction, nor do they support adaptive recursion.

More recently, recursive transformer architectures have been explored for vision tasks. TRM-ViT \cite{sabri2026trm} introduces a lightweight recursive Vision Transformer for melanoma detection, demonstrating that parameter-efficient recursive computation can be effective in medical imaging. However, its design remains task-specific and does not explicitly model a dual-state recursion or incorporate deep supervision for progressive prediction refinement.

ViTRM builds upon these ideas by maintaining a compact latent reasoning state $\mathbf{z}$ and a task-specific output $\mathbf{y}$, both recursively refined through shared-weight updates. Crucially, optional cross-attention from $\mathbf{z}$ to image tokens $\mathbf{x}$ enables repeated spatial querying, bridging object-centric latent refinement with task-driven visual prediction.

\subsection{Deep Equilibrium and Fixed-Point Methods}

Deep Equilibrium Models (DEQ) \cite{bai2019deep, krantz2013implicit} offer an alternative perspective on depth by solving for fixed points of implicit functions. Instead of stacking many layers, DEQ defines the output as the fixed point $\mathbf{z}^* = f_\theta(\mathbf{z}^*, \mathbf{x})$ and solves for $\mathbf{z}^*$ via root-finding algorithms. 
TorchDEQ \cite{geng2023torchdeq} provides a modular library for implementing such models, enabling easier adoption across domains. 
Although DEQ and recursive reasoning models such as TRM \cite{jolicoeurmartineau2025recursive, wang2025hierarchicalreasoningmodel} both rely on repeated function application, their objectives differ fundamentally.

Vision tasks often involve dense outputs 
where iterative refinement naturally improves predictions but strict convergence to a fixed point may be unnecessary. By embracing TRM's philosophy of unrolled recursion with deep supervision, ViTRM gains flexibility to adapt recursion depth at test time while maintaining stable gradients through full backpropagation rather than relying on 1-step gradient approximations \cite{bai2019deep} that assume convergence.
\subsection{Adaptive Computation and Dynamic Inference}
Adaptive computation, through Adaptive Computation Time (ACT) \cite{graves2016adaptive}, seeks to distribute computational resources based on the difficulty of the input, considering latency, energy, and accuracy. 
Although elegant, ACT may be highly sensitive to hyperparameters such as halting thresholds and time penalties. Early-exit methods \cite{teerapittayanon2016branchynet, kaya2019shallow} add intermediate classifiers to allow predictions at various depths. 
Nevertheless, these methods still require auxiliary classifiers and confidence thresholds.

ViTRM takes a simpler 
approach: During inference, the number of recursive calls can be directly controlled, allowing a straightforward trade-off between latency and accuracy. 
Compared to ACT, ViTRM employs a light-weight halting function inspired by TRM \cite{jolicoeurmartineau2025recursive}. Instead of summing ponder costs or using reinforcement learning (e.g., Q-learning \cite{wang2025hierarchicalreasoningmodel}), it predicts a binary halting probability 
which enables early stopping during training without incurring extra forward passes. 
\subsection{Recursive Reasoning Models: From HRM to TRM and Beyond}

The most direct precursors to ViTRM are the Hierarchical Reasoning Model (HRM) \cite{wang2025hierarchicalreasoningmodel} and Tiny Recursion Model (TRM) \cite{jolicoeurmartineau2025recursive}, both designed for discrete reasoning tasks like Sudoku, maze solving, and ARC-AGI puzzles.

HRM introduced recursive reasoning with two networks operating at different frequencies: a low-level network $f_L$ applied multiple times and a high-level network $f_H$ applied less frequently, maintaining two latent features $\mathbf{z}_L$ and $\mathbf{z}_H$. 


TRM \cite{jolicoeurmartineau2025recursive} dramatically simplified HRM while improving generalization. It replaced HRM's two 4-layer networks with a single 2-layer network, reinterpreted $\mathbf{z}_H$ as the current embedded answer $\mathbf{y}$ and $\mathbf{z}_L$ as a latent reasoning state $\mathbf{z}$, removed the need for fixed-point assumptions by backpropagating through full recursion cycles, and simplified halting to avoid extra forward passes.


TRM demonstrated that "less is more": tiny networks with deep recursion outperformed larger networks, achieving 87\% accuracy on Sudoku-Extreme compared to HRM's 55\%, despite using 7M parameters versus 27M. The core insight is that weight sharing combined with deep supervision acts as a powerful regularizer, preventing overfitting on small datasets while enabling effective depths far exceeding what standard training permits.\\

However, prior recursive reasoning models target low-dimensional symbolic states, whereas vision requires iterative reasoning over high-dimensional spatial tokens, motivating a recursion mechanism that can query image features.

ViTRM bridges these gaps by extending TRM's recursive reasoning framework to vision. Specifically, ViTRM introduces:

\begin{itemize}
\item \textbf{Cross-attention from $\mathbf{z}$ to image tokens $\mathbf{x}$}: Enables the latent reasoning state to iteratively query spatial features, essential for tasks like semantic segmentation, object detection, and dense prediction where the model must attend to specific image regions during reasoning.

\item \textbf{Dual-state recursion for vision}: Maintains both a task output $\mathbf{y}$ (e.g. class logits) and a latent reasoning state $\mathbf{z}$, recursively refining both via a tiny shared-weight block.
\item \textbf{Deep supervision with simple halting}: Trains the model to improve predictions at every supervision step while enabling efficient training through learned early stopping and removing output and reasoning states from gradient computation  between supervision steps,   without extra forward passes.
\item \textbf{Architectural flexibility}: Supports both attention-based and MLP-based token mixing, adapting to task-specific inductive biases (e.g., MLP-Mixer-style layers for fixed-resolution tasks, self-attention for variable-length sequences).
\end{itemize}

By combining the parameter efficiency and deep recursion of TRM with architectural adaptations for spatial reasoning, ViTRM offers a principled approach to vision tasks that naturally scales with test-time compute. Unlike LLM-oriented recursive reasoning, which operates over discrete token sequences and autoregressive generation, ViTRM operates over dense spatial representations, refining pixel-aligned outputs through iterative cross-attention.


In summary, ViTRM adapts TRM-style recursive reasoning to vision by iteratively refining a latent state and outputs over spatial tokens.

\section{Vision Tiny Recursion Model}\label{sec3}
\begin{figure}[H]
  \centering
  \includegraphics[width=1.1\linewidth]{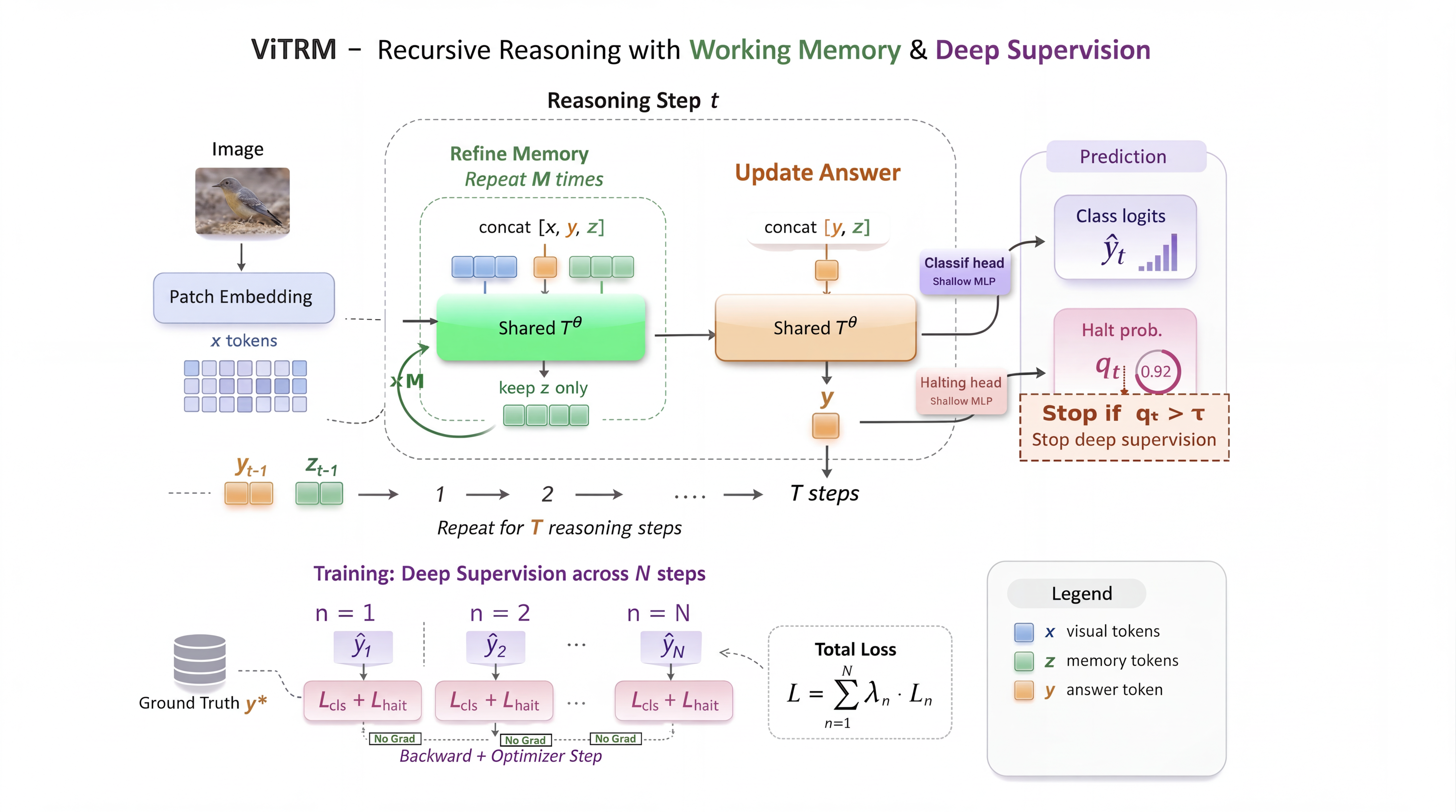}
  \caption{
  \textbf{ViTRM: Recursive Reasoning with Working Memory and Deep Supervision.}
  \textit{Top:} At each reasoning step $t$, the model alternates between two phases sharing the same transformer weights $\theta$. 
  In the \textbf{Refine Memory} phase, the concatenation of visual tokens $x$, answer token $y$, and memory tokens $z$ is processed by the shared transformer $T^\theta$ for $M$ iterations, retaining only the updated memory $z$. 
  In the \textbf{Update Answer} phase, the concatenation of $y$ and $z$ is fed to $T^\theta$ to produce a new answer token $y$. This process is repeated $T$ times and then passed to two shallow MLP heads: a \textbf{classification head} producing class logits $\hat{y}_t$, and a \textbf{halting head} producing a halting probability $q_t$. Inference stops when $q_t > \tau$.
  \textit{Bottom:} During training, \textbf{deep supervision} is applied at each of the $N$ reasoning steps. At step $n=1,\cdots, N$,  leveraging  previous $z$ and $y$ (detached from gradients), the predicted output $\hat{y}_n$ is supervised with a combined loss $L_n = L^n_\text{cls} + L^n_\text{halt}$, using stop-gradient to prevent interference between steps. 
}
  \label{fig1}
\end{figure}

We present the \textbf{Vision Tiny Recursion Model (ViTRM)}, a
parameter-efficient architecture for image classification that extends
Tiny Recursion Models~\cite{jolicoeurmartineau2025recursive} to visual
domains. Rather than increasing model capacity through additional layers,
ViTRM applies two small shared Transformer blocks recursively over $T$
steps, progressively refining two internal states through iterative
interaction with fixed image features: a latent memory
$\mathbf{z} \in \mathbb{R}^{K \times d}$, where $d$ is the token
embedding dimension and $K$ is a fixed number of latent tokens, and a
prediction token $\mathbf{y} \in \mathbb{R}^{1 \times d}$, a single
vector that distils the latent memory into a compact class hypothesis.
Two linear heads read from $\mathbf{y}$ at each step: a classification
head $W_c \in \mathbb{R}^{C \times d}$ producing logits over $C$ classes,
and a halting head $W_h \in \mathbb{R}^{1 \times d}$ estimating whether
the current prediction is already reliable to stop the recursive process. This design achieves deep
reasoning at inference time while keeping the parameter count small, as
all recursive steps share the same weights. Figure~\ref{fig1} illustrates
the full architecture.

\subsection{Patch Embedding}

Following ViT~\cite{dosovitskiy2021imageworth16x16words}, the input image
$\mathcal{I} \in \mathbb{R}^{H \times W \times 3}$ is divided into
non-overlapping $P \times P$ patches, yielding $L_x = HW / P^2$ tokens.
Each patch is linearly projected into a $d$-dimensional embedding and
summed with a learnable positional embedding, producing the image token
sequence
\begin{equation}
  \mathbf{x} = f_E(\mathcal{I}) \;\in \mathbb{R}^{L_x \times d},
\end{equation}
which remains \emph{fixed} throughout all subsequent recursion steps.
We use a simple linear projection for $f_E$ to minimize parameters and
delegate all feature refinement to the recursive blocks.

\subsection{Recurrent States}

ViTRM maintains two recurrent states, both initialized from learned
embeddings and updated at every reasoning step:
\begin{itemize}
  \item \textbf{Prediction token} $y^t \in \mathbb{R}^{d}$: a single
    vector analogous to the \texttt{[CLS]} token in standard Vision
    Transformers, representing the model's current class hypothesis.
  \item \textbf{Latent memory} $z^t \in \mathbb{R}^{K \times d}$: a
    set of $K$ free tokens, where $K$ is a fixed hyperparameter
    \emph{independent of the number of image patches} $L_x$.
    Rather than maintaining a patch-aligned representation, ViTRM
    compresses visual information into a compact bottleneck of $K \ll L_x$
    tokens, reducing the computational cost of each inner iteration and
    encouraging the latent state to learn task-relevant abstractions.
\end{itemize}

The separation between $y^t$ and $z^t$ is deliberate: the latent memory
absorbs and refines spatial information from the image patches, while the
prediction token distills this into a compact task-relevant representation.
During training, the model is unrolled for up to $N$ \textbf{supervision
steps}. At each step $n \in \{1, \ldots, N\}$, both states are refined
through $T$ recursion steps initialized from the detached states (no gradient) of the
previous step; the update rules are described below.

\subsection{Recursive Update Rules}

ViTRM uses a shared single Transformer blocks $\mathcal{T}^\theta$   that  refines $z$ by attending jointly
over the image tokens, the prediction token, and the current latent state.
It also  updates $y$ from the refined
latent state and the current prediction token, with no direct access to the image patches. It comprises 
 multi-head self-attention (that fosters contextual reasoning), a GELU shallow feed-forward network, layer normalization,
and a residual connection.

At each recursion step $t \in \{1, \ldots, T\}$, the latent memory is
first refined over $M$ inner iterations. Starting from $z^{t,0} = z^t$,
each inner iteration is:
\begin{equation}
  z^{t,k+1}
    =\mathcal{T}^\theta\!\bigl([\mathbf{x},\, y^t,\, z^{t,k}]\bigr)_{[2:]}
  \,, \quad k = 0,\dots,M{-}1,
  \label{eq:latent}
\end{equation}
where $[\cdot]_{[2:]}$ denotes the last $K$ output tokens, and the input
sequence has length $L_x + 1 + K$. After the inner loop,
$z^{t+1} = z^{t,M}$. The prediction token is then updated from the
refined memory alone:
\begin{equation}
  y^{t+1}
    = \mathcal{T}^\theta\!\bigl([y^t,\, z^{t+1}]\bigr)_{[1]},
  \label{eq:pred}
\end{equation}
where $[\cdot]_{[1]}$ denotes the first output token. Excluding
$\mathbf{x}$ from Eq.~\eqref{eq:pred} ensures that visual information
reaches $y$ exclusively through the abstracted memory $z$, contributing
to training stability under truncated gradients and forcing the latent
block to produce informative representations.

\subsection{Classification and Halting Heads}

Two linear heads read from the prediction token $y_T^{(n)} \in \mathbb{R}^d$, the state of $y$ after $T$ recursion steps at supervision step $n$,
and produce a class prediction and a confidence estimate:
\begin{equation}
  \ell^{(n)} = W_c\,y_T^{(n)} \in \mathbb{R}^C,
  \qquad
  p^{(n)}   = W_h\,y_T^{(n)} \in \mathbb{R},
  \label{eq:heads}
\end{equation}
where $W_c \in \mathbb{R}^{C \times d}$ produces logits over $C$ classes
and $W_h \in \mathbb{R}^{1 \times d}$ is the halting head, whose sigmoid
output $q^{(n)} = \sigma(p^{(n)}) \in [0,1]$ estimates the probability
that the current prediction is already correct.

\subsection{Training}

At each supervision step $n$, the model performs $T$ recursion steps via
Eqs.~\eqref{eq:latent}-\eqref{eq:pred}, starting from the detached
states of the previous step, and the heads in Eq.~\eqref{eq:heads} are
applied to produce $\ell^{(n)}$ and $q^{(n)}$. A gradient update is
applied immediately after each step, before states $(y^{(n)}, z^{(n)})$
detached from the computation graph before being passed to step 
$n+1$, yielding $N$ independent weight
updates per batch. This resembles truncated backpropagation through
time~\citep{aicher2020adaptively}, keeping gradient computation tractable
while preserving a warm initialization from the previous step. We find
empirically that this per-step update scheme outperforms the strict TRM
training procedure~\cite{jolicoeurmartineau2025recursive}, which performs
$T{-}1$ gradient-free passes followed by a single backpropagation step; we omit this comparison 
from our results as it does not bear on the main contributions of this work.

The per-step loss combines a classification term and a halting term:
\begin{equation}
  \mathcal{L}^{(n)} =
    \mathcal{L}_{\mathrm{CE}}\!\left(\ell^{(n)},\, y^*\right)
    +
    \mathcal{L}_{\mathrm{BCE}}\!\left(q^{(n)},\,
      \mathbf{1}\!\left\{\hat{y}^{(n)} = y^*\right\}\right),
  \qquad
  \hat{y}^{(n)} = \arg\max\,\ell^{(n)},
  \label{eq:loss}
\end{equation}
where $y^* \in \{1,\ldots,C\}$ is the ground-truth label. The halting
target is $1$ when the current prediction is correct and $0$ otherwise,
training the model to recognize when further recursion is unnecessary.
Unrolling terminates early when the mean halting probability across the
batch of size $B$ exceeds a threshold $\tau \in(0,1)$ ($\tau=0.5$ in our experiments):
\begin{equation}
  q=\frac{1}{B}\sum_{i=1}^{B} q_i^{(n)} > \tau.
  \label{eq:halt}
\end{equation}
At inference, the model runs exactly $T$ recursion steps without the
halting condition, returning the logits of the final step.

\section{Experiments}
\label{sec4}
We evaluate ViTRM against established CNN and Vision Transformer baselines on image classification benchmarks to assess whether recursive computation can achieve competitive accuracy with significantly fewer parameters.
\paragraph{Vision Transformers (ViT).}
We evaluate three variants of Vision Transformer \cite{dosovitskiy2021imageworth16x16words}: ViT-Small (16.9M params), ViT-Base (12 layers, 85.1M), and ViT-Large (302M). Since ViTRM also operates on patch embeddings with self-attention, these models provide a direct architectural comparison across different parameter scales, allowing us to assess whether recursive weight sharing can match deeper non-shared Transformers.

\paragraph{ResNets.}
We additionally compare against ResNet variants \cite{he2016deep}: ResNet-18 (11.2M), ResNet-34 (21.3M), and ResNet-50 (23.5–23.7M). ResNets represent convolutional architectures with strong locality biases, providing a complementary baseline to contextualize ViTRM’s performance.
\subsection{Experimental Setup}

We conduct experiments on two standard image classification benchmarks:
\begin{itemize}
    \item \textbf{CIFAR-10} \citep{NIPS2012_c399862d}: 60,000 images (50,000 train, 10,000 test) across 10 classes at $32 \times 32$ resolution.
    \item \textbf{CIFAR-100} \citep{russakovsky2015imagenet}: 60,000 images with 100 fine-grained classes at the same resolution.
\end{itemize}

All images are normalized using mean and standard deviation values. We apply a standard augmentation pipeline consisting of random augmentations \cite{cubuk2019randaugmentpracticalautomateddata}, Mixup \cite{zhang2018mixupempiricalriskminimization}, and CutMix \cite{yun2019cutmixregularizationstrategytrain}  during training.

We report Top-1 classification accuracy on the test set. All models are trained from scratch without ImageNet pre-training to ensure a fair comparison of architectural capacity rather than transfer learning effects. We use the standard test set for validation-based early stopping, selecting the checkpoint with the best validation accuracy. Early stopping is triggered when validation accuracy does not improve across 10 consecutive epochs.

To study the sensitivity of each architecture to optimization dynamics, we train all models across five batch sizes: $64$, $128$, $256$, $512$, and 1024. This sweep reveals how models behave under different gradient noise regimes and whether ViTRM's recursive structure confers any robustness to hyperparameter choices.



\subsection{Implementation Details}
All experiments were performed on a single NVIDIA GH200 GPU with 120GB of memory. All models are trained using AdamW \cite{loshchilov2019decoupledweightdecayregularization} with an initial learning rate of $3 \times 10^{-4}$, weight decay of $0.05$, and a cosine annealing schedule with linear warmup over the first $5\%$ of training epochs. ViT and ResNet baselines are trained for 300 epochs. ViTRM is allotted up to 1000 epochs to account for its slower per-epoch convergence characteristic of recursive models with deep supervision. Importantly, ViTRM's tiny encoder (3.6M parameters) results in significantly faster wall-clock time per epoch compared to ViT-Base or ViT-Large, making extended training computationally tractable. Early stopping ensures that ViTRM does not overtrain when convergence is reached before the maximum epoch budget.

All Transformer-based models (ViT variants and ViTRM) use a patch size of $4 \times 4$ to accommodate the $32 \times 32$ CIFAR resolution, yielding $8 \times 8 = 64$ patches per image.

ViTRM applies a tiny $3$-layer encoder recursively for $N_{\text{rec}}$ steps, as described in Section~\ref{sec3}. We use the following hyperparameters:
\begin{itemize}
    \item \textbf{Recursive steps}: $n_{\text{latent\_steps}} = 3$
    \item \textbf{Supervision frequency}: $N_{\text{supervision}} = 1$
    \item \textbf{Deep supervision depth}: $T_{\text{deep}} = 1$
\end{itemize}

As in TRM, We apply EMA to model weights during training with a decay rate of 0.999. This stabilizes training for recursive architectures where the same weights are applied multiple times per forward pass.

\subsection{Main Results}
Tables~\ref{tab1} and~\ref{tab2} present Top-1 test accuracy across all models and batch sizes for CIFAR-10 and CIFAR-100, respectively.

\begin{table*}[t]
\centering
\caption{Top-1 test accuracy (\%) on \textbf{CIFAR-10} across batch sizes. ViTRM achieves competitive accuracy with $\sim$3.6M parameters-4.7$\times$ fewer than ViT-Small and 23$\times$ fewer than ViT-Base. Best result per column in \textbf{bold}; best ViTRM result \underline{underlined}.}
\label{tab1}
\begin{tabular}{lrrrrrr}
\toprule
\textbf{Model} & \textbf{\#Params} & \textbf{BS=64} & \textbf{BS=128} & \textbf{BS=256} & \textbf{BS=512} & \textbf{BS=1024} \\
\midrule
ViTRM (ours)  & 3.6M   & 92.6 & \underline{93.1} & 92.8 & 92.1 & 91.2 \\
\midrule
ViT-Small     & 16.9M  & \textbf{94.0} & \textbf{93.9} & 93.3 & 92.8 & 91.1 \\
ViT-Base      & 85.1M  & 93.7 & 93.8 & \textbf{94.3} & \textbf{94.1} & \textbf{94.0} \\
ViT-Large     & 302.4M & 80.8 & 73.4 & 82.0 & 83.2 & OOM \\
\midrule
ResNet-18     & 11.2M  & 90.5 & 89.7 & 88.8 & 87.8 & 86.1 \\
ResNet-34     & 21.3M  & 90.8 & 90.1 & 89.7 & 87.9 & 86.3 \\
ResNet-50     & 23.5M  & 92.0 & 90.9 & 89.7 & 87.9 & 84.2 \\
\bottomrule
\end{tabular}
\end{table*}

\begin{table*}[t]
\centering
\caption{Top-1 test accuracy (\%) on \textbf{CIFAR-100} across batch sizes. ViTRM maintains strong performance on the more challenging 100-class task while using a fraction of baseline parameters.}
\label{tab2}
\begin{tabular}{lrrrrrr}
\toprule
\textbf{Model} & \textbf{\#Params} & \textbf{BS=64} & \textbf{BS=128} & \textbf{BS=256} & \textbf{BS=512} & \textbf{BS=1024} \\
\midrule
ViTRM (ours)  & 3.7M   & 71.8 & \underline{72.1} & 72.0 & 71.3 & 69.8 \\
\midrule
ViT-Small     & 16.9M  & \textbf{76.1} & 75.2 & 74.8 & 72.7 & 71.2 \\
ViT-Base      & 85.2M  & 72.4 & 74.6 & 75.4 & \textbf{76.4} & \textbf{75.1} \\
ViT-Large     & 302.5M & 61.1 & 58.9 & 61.7 & 64.7 & OOM \\
\midrule
ResNet-18     & 11.2M  & 64.6 & 63.4 & 62.1 & 60.5 & 58.2 \\
ResNet-34     & 21.3M  & 65.4 & 63.8 & 62.0 & 58.8 & 56.7 \\
ResNet-50     & 23.7M  & 63.8 & 63.4 & 61.5 & 58.0 & 54.5 \\
\bottomrule
\end{tabular}
\end{table*}

\subsubsection{ViTRM vs.\ Vision Transformers}
On CIFAR-10, ViTRM achieves 93.1\% accuracy (BS=128), within 0.9 percentage points of ViT-Small (94\%) while using \textbf{4.7$\times$ fewer parameters}. Compared to ViT-Base, ViTRM trails by only 1.2 points at the optimal batch size despite having \textbf{23$\times$ fewer parameters}.

ViT-Large consistently underperforms smaller variants on both datasets (peaking at 83.2\% on CIFAR-10 and 64.7\% on CIFAR-100), indicating severe overfitting when training from scratch on limited data without pre-training. This highlights that scaling depth and width without pre-training or strong regularization is counterproductive on small-scale benchmarks.

On CIFAR-100, ViTRM (72.1\%) narrows the gap with ViT-Small (76.1\%) to 4 percentage points and matches ViT-Base at smaller batch sizes (72.4\% at BS=64). The consistent performance across batch sizes suggests that ViTRM's recursive weight-sharing provides implicit regularization beneficial for fine-grained classification.

\subsubsection{ViTRM vs.\ ResNets}
ViTRM outperforms all ResNet variants on both datasets across all batch sizes. On CIFAR-10, ViTRM (93.1\%) exceeds ResNet-50 (92.0\%) by 1.1 points while using \textbf{6.5$\times$ fewer parameters}. The gap widens on CIFAR-100: ViTRM achieves 72.1\% versus ResNet-50's 63.8\%, a margin of \textbf{8.3 percentage points}.

ResNets exhibit pronounced sensitivity to batch size, with accuracy degrading substantially at larger batch sizes (e.g., ResNet-50 drops from 92.0\% to 84.2\% on CIFAR-10 as batch size increases from 64 to 1024). ViTRM shows more graceful degradation (93.1\% to 91.2\%), suggesting that its recursive structure is more robust to optimization hyperparameters.

\subsubsection{Batch Size Sensitivity}
All models exhibit some degree of batch size sensitivity, with smaller batch sizes generally yielding better results. However, the magnitude of degradation varies:
\begin{itemize}
    \item \textbf{ViT-Base} is remarkably stable (93.7 - 94.3\% on CIFAR-10), benefiting from its large capacity.
    \item \textbf{ViTRM} exhibits strong robustness to large batch sizes, with only a minor drop in accuracy (93.1\% $\rightarrow$ 91.2\% on CIFAR-10), remaining highly competitive even at BS=1024.
    \item \textbf{ResNets} degrade sharply (ResNet-50: 92.0\% $\rightarrow$ 84.2\%), suggesting that convolutional architectures trained from scratch are more sensitive to batch size on these benchmarks.
\end{itemize}

The optimal batch size for ViTRM is 128, achieving peak accuracy on both datasets. This is consistent with observations in TRM\cite{jolicoeurmartineau2025recursive} that recursive models benefit from moderate gradient noise during training.

\subsection{Ablation Studies on Supervision and Latent Reasoning Depths}
\subsubsection{Supervision Depth}

\begin{figure}[ht!]
  \centering
  \includegraphics[width=0.65\linewidth]{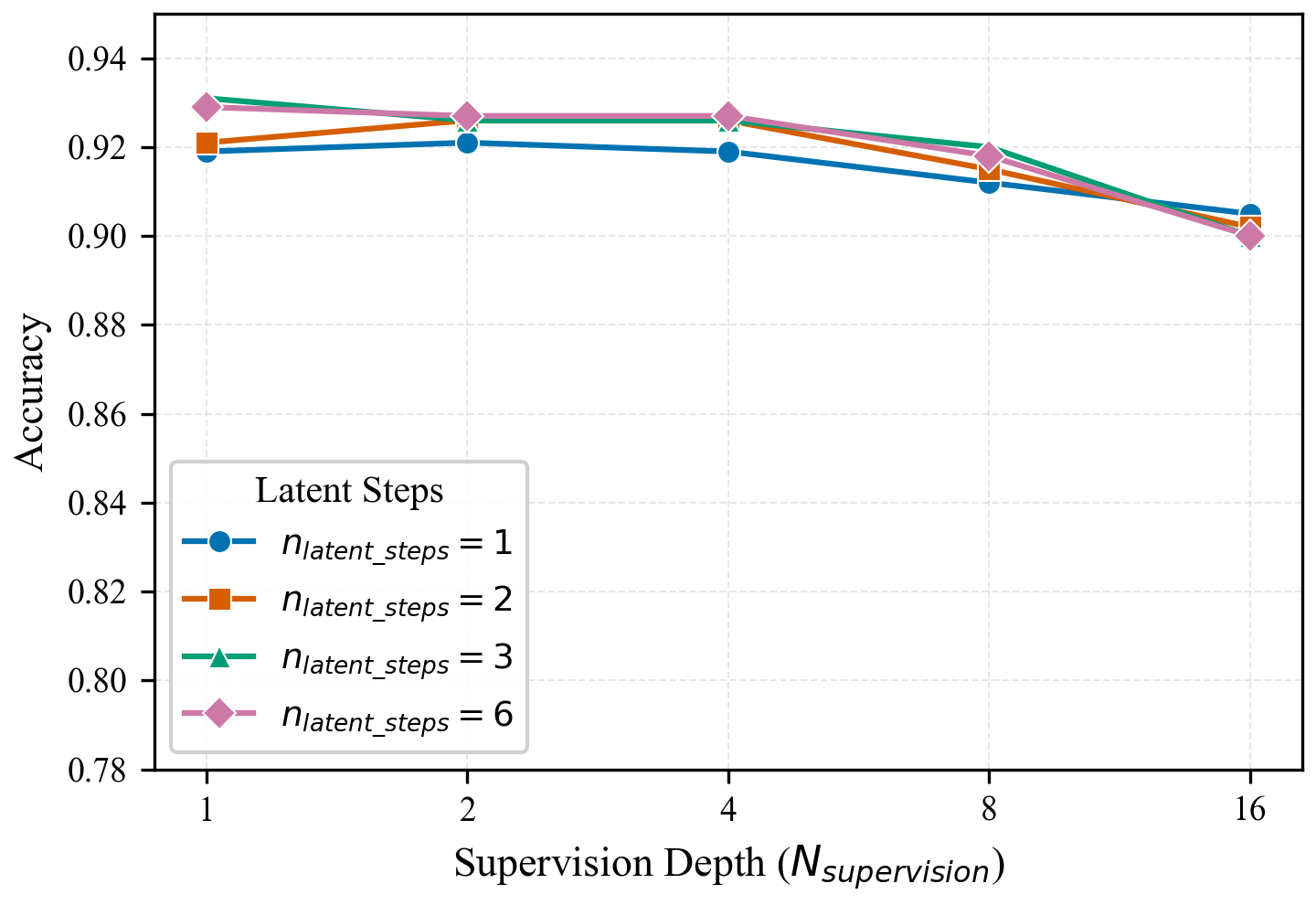}
  \caption{Ablation results for supervision depth $N_{\text{supervision}}$ on CIFAR-10.}
  \label{fig2}
\end{figure}

We investigate the effect of supervision depth $N_{\text{supervision}}$ on model performance by varying this hyperparameter across values $\{1, 2, 4, 8, 16\}$ while holding other factors constant. All experiments are conducted on CIFAR-10 with batch size 128. Figure~\ref{fig2} presents the results across multiple latent reasoning depths.
Our findings reveal a consistent pattern: performance degrades as supervision depth increases beyond minimal values. At the default latent depth ($n_{\text{latent}}=3$), the model achieves 93.1\% top-1 accuracy with $N_{\text{supervision}}=1$, which decreases to 92.6\% at $N_{\text{supervision}}=2$ and further declines to 90\% at $N_{\text{supervision}}=16$. This trend holds across all tested latent depths, with the cofiguration $N_{\text{supervision}}=16$ and  $n_{\text{latent\_steps}}=6$, yielding the lowest accuracy of 90\%.

We hypothesize that an excessive supervision depth leads primarily to overfitting. By imposing supervision too densely, the model may become overly constrained to fit training-specific signals, which can reduce its ability to learn flexible and generalizable intermediate representations that are truly optimal for the task. These results justify our default choice of $N_{\text{supervision}}=1$, which consistently achieves the best or near-best performance across all configurations while minimizing computational overhead from auxiliary supervision losses.

\subsubsection{Latent Reasoning Depth}

\begin{figure}[ht!]
  \centering
  \includegraphics[width=0.65\linewidth]{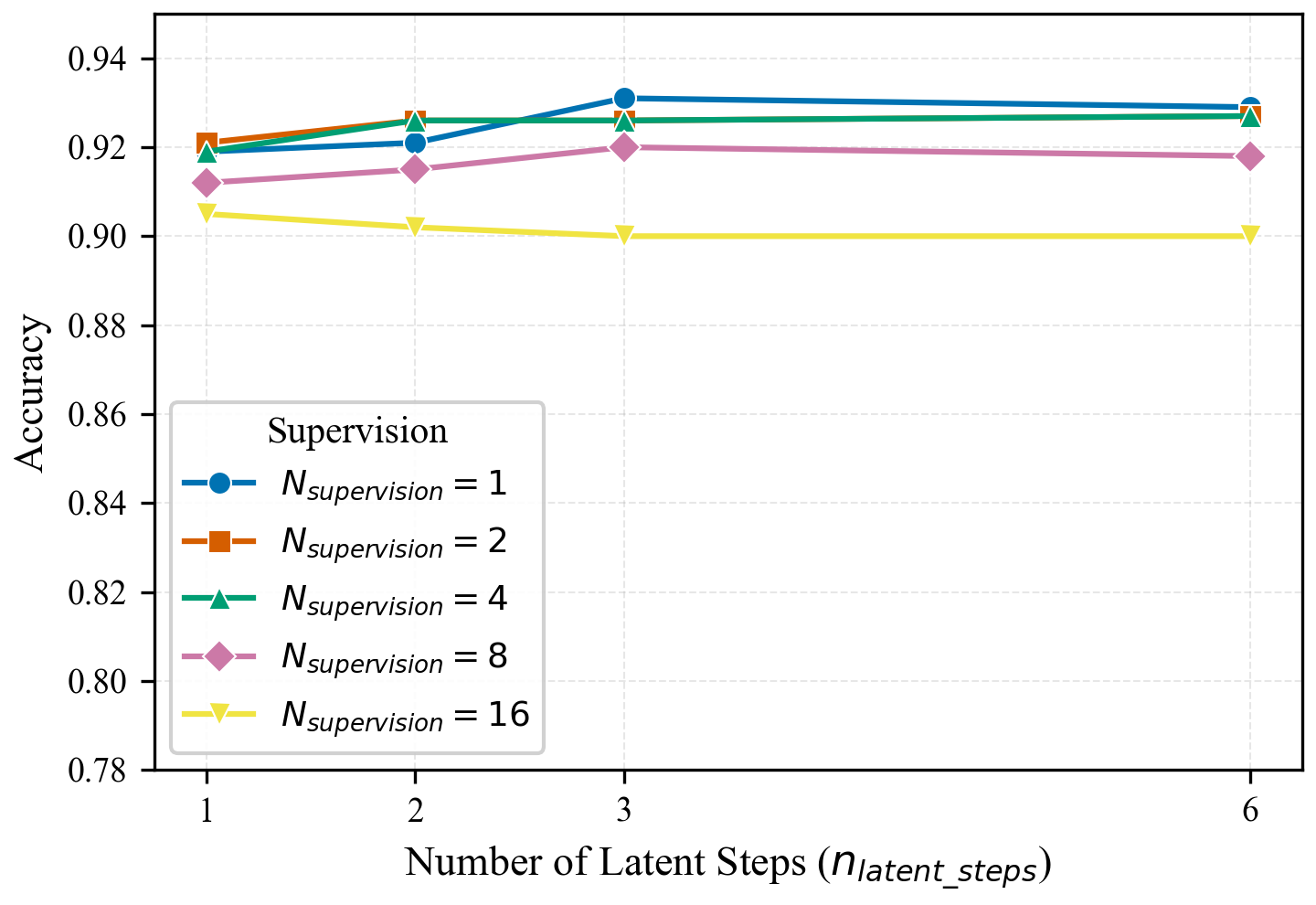}
  \caption{Ablation results for reasoning depth $n_{\text{latent\_steps}}$ on CIFAR-10.}
  \label{fig3}
\end{figure}

We examine how the number of latent reasoning steps $n_{\text{latent\_steps}}$ affects model accuracy by testing values $\{1, 2, 3, 6\}$ across different supervision depths. Results are shown in Figure~\ref{fig3}.

At the default supervision depth ($N_{\text{supervision}}=1$), increasing latent reasoning depth improves performance overall, with accuracy rising from 91.9\% at $n_{\text{latent\_steps}}=1$ to 92.9\% at $n_{\text{latent\_steps}}=6$, corresponding to a net gain of 1.0 percentage point. However, the improvement is not monotonic: accuracy increases slightly to 92.1\% at $n_{\text{latent\_steps}}=2$, peaks at 93.1\% at $n_{\text{latent\_steps}}=3$, and then decreases slightly to 92.9\% at $n_{\text{latent\_steps}}=6$. This suggests that moderate latent depth yields the best performance, while further increasing depth provides no additional benefit.

Notably, the effect of increased latent depth depends on the supervision depth. For moderate supervision levels ($N_{\text{supervision}} \in \{2,4,8\}$), performance saturates rapidly, with $n_{\text{latent\_steps}}=3$ already matching deeper configurations. By contrast, under stronger supervision ($N_{\text{supervision}}=16$), larger latent depth is not only unhelpful but slightly harmful, as accuracy decreases from 90.5\% to 90.0\% when $n_{\text{latent\_steps}}$ increases from 1 to 6. This indicates that, in this regime, additional latent steps add computation without improving accuracy.

These results suggest that latent reasoning depth and supervision depth interact non-trivially. Our default configuration of $N_{\text{supervision}}=1$ and $n_{\text{latent\_steps}}=3$ provides a favorable trade-off between accuracy and computational cost, achieving 93.1\% accuracy while avoiding the diminishing returns observed at higher depths and maintaining robustness across supervision settings.

\section{Conclusion}
We have presented ViTRM, a parameter-efficient vision architecture that challenges the prevailing ``deeper and larger is better'' paradigm in computer vision. By coupling the patch-based tokenization of Vision Transformers with the recursive state refinement mechanism of Tiny Recursive Models, ViTRM demonstrates that architectural depth can be effectively replaced by iterative computation with shared weights. Our model maintains only 3.6M parameters yet achieves accuracy competitive with ViTs and substantially outperforms ResNet variants on both CIFAR-10 and CIFAR-100, establishing recursive computation as a principled approach to parameter-efficient visual recognition.

Our ablation studies reveal important design principles for recursive vision models. We find that minimal supervision depth ($N_{\text{supervision}}{=}1$) yields optimal performance, with excessive supervision leading to overfitting and degraded generalization. Similarly, latent reasoning depth exhibits diminishing returns beyond $n_{\text{latent\_steps}}{=}3$, suggesting that a moderate number of refinement steps suffices for the benchmarks studied. These findings are in agreement with the recent results reported in \cite{royeazar2026tinyrecursivemodelsarcagi1}, and indicate that the regularization effect of weight sharing, combined with deep supervision at sparse intervals, prevents overfitting even under extended training schedules.

This study provides a strong foundation for several future research directions. ViTRM has been validated primarily on moderate-resolution classification benchmarks; scaling to high-resolution inputs and dense prediction tasks such as semantic segmentation and object detection remains to be explored. The interaction between recursive depth and task complexity also merits deeper theoretical analysis. Looking forward, ViTRM's cross-attention mechanism from latent states to image tokens opens promising directions for dense reasoning tasks, while its lightweight halting mechanism may enable adaptive test-time computation. We believe recursive architectures represent a compelling path toward accessible, deployable vision models that maintain strong performance without prohibitive resource requirements.


\medskip




\bibliographystyle{plain}
\bibliography{main}


\end{document}